\documentclass[journal]{IEEEtran}

\usepackage{tikz, tikzpagenodes}
\usepackage{pgfplots}
\usepackage{caption}
\usepackage{float}
\pgfplotsset{compat=1.18} 

\usepackage{booktabs}
\usepackage{amsmath}
\usepackage{comment}
\usepackage{longtable}
\usepackage{dsfont}
\usepackage{amsfonts}
\usepackage{subcaption}

\usepackage{titlesec}
\titlespacing*{\section}{0pt}{10pt plus 2pt minus 2pt}{5pt plus 2pt minus 2pt}

\titlespacing*{\subsection}{0pt}{10pt plus 4pt minus 2pt}{5pt plus 2pt minus 2pt}

\usepackage{url}

\begin{document}

\title{PGNAA Spectral Classification of Aluminium and Copper Alloys with Machine Learning}

\author{Henrik Folz$^*$,
        Joshua Henjes$^*$,
        Annika Heuer$^*$,
        Joscha Lahl$^*$,
        Philipp Olfert$^*$,
        Bjarne Seen$^*$,
        Sebastian Stabenau$^*$, 
        Kai Krycki,
        Markus Lange-Hegermann,
        Helmand Shayan

\thanks{All first-contributors are with the Department
of Electrical Engineering and Computer Science, OWL University of Applied Sciences and Arts, Lemgo, 
32657 Germany.}%
\thanks{Kai Krycki is with the Department of Aerospace Engineering, FH Aachen - University of Applied Sciences, Aachen, 52064 Germany.}%
\thanks{Helmand Shayan and Markus Lange-Hegermann are with the inIT - Institute Industrial IT, OWL University of Applied Sciences and Arts, Lemgo, 32657 Germany.\\}
\thanks{* equal contribution, alphabetical order}%
}

\maketitle

\begin{abstract}
In this paper, we explore the optimization of metal recycling with a focus on real-time differentiation between alloys of copper and aluminium. Spectral data, obtained through Prompt Gamma Neutron Activation Analysis (PGNAA), is utilized for classification. The study compares data from two detectors, cerium bromide (CeBr$_{3}$) and high purity germanium (HPGe), considering their energy resolution and sensitivity. We test various data generation, preprocessing, and classification methods, with Maximum Likelihood Classifier (MLC) and Conditional Variational Autoencoder (CVAE) yielding the best results. The study also highlights the impact of different detector types on classification accuracy, with CeBr$_{3}$ excelling in short measurement times and HPGe performing better in longer durations. The findings suggest the importance of selecting the appropriate detector and methodology based on specific application requirements.
\end{abstract}


\section{Introduction}
\IEEEPARstart{I}{n} 2019, humankind mined 62.9 million tonnes of aluminium and 20.7 million tonnes of copper in the world~\cite{Metal.mining}. The worldwide recycling rate of End of Life products from copper is 45\%~\cite{HENCKENS2020121460}, while the rate for aluminium is 70\%~\cite{Recycling.Aluminium}. With recycling of aluminium 95\% of energy and 90\% of {\ensuremath{\mathrm{CO_2}}} can be saved compared with mining of aluminium~\cite{Recycling.Aluminium}~\cite{Recycling.Aluminium2}.
This shows the increasing importance of developing efficient recycling processes. 

For recycling of scrap metal into \emph{high-quality} alloys, one needs to know the atomic composition of the scrap in detail.
Currently, there are mainly sensor methods that can classify scrap metal using traditional chemical processes or Laser Induced Breakdown Spectroscopy (LIBS)~\cite{park20223d}.
Both methods only consider the analysis of the scrap metal surface instead of integrally measuring the scrap metal composition and chemical methods even destroy the material.

\begin{figure}[t]
    \centering
    \begin{subfigure}[b]{0.45\textwidth}
        \includegraphics[width=\textwidth]{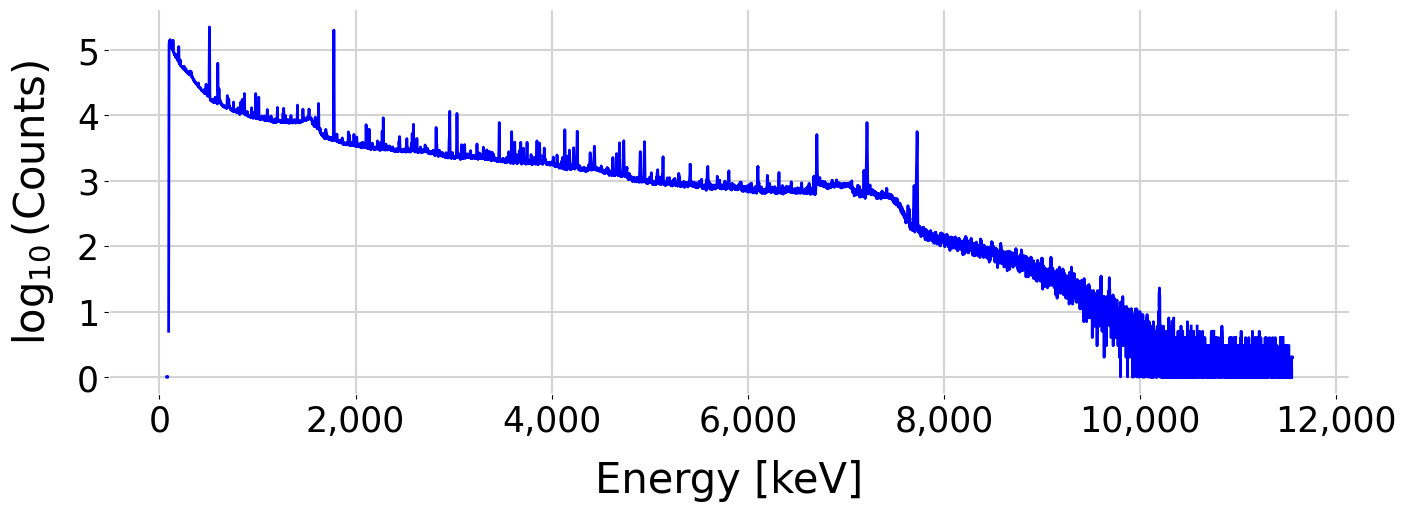}
        \caption{\emph{Long-term measurement (1h)}}
        \label{fig:plot_a}
    \end{subfigure}
    \hfill
    \begin{subfigure}[b]{0.45\textwidth}
        \includegraphics[width=\textwidth]{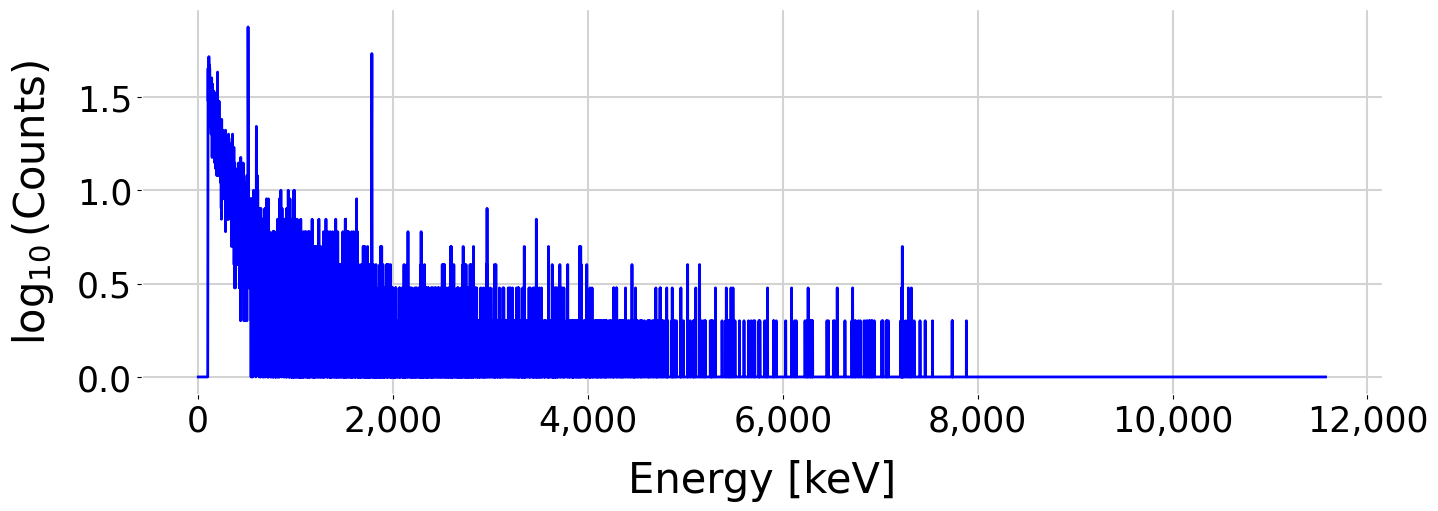}
        \caption{\emph{Short-term measurement (1s)}}
        \label{fig:plot_b}
    \end{subfigure}
    \caption{\emph{Example prompt gamma spectra of the gamma quants of an aluminium alloy measured with a HPGe detector as basis for classification. The long-term measurement of 1h in (\subref{fig:plot_a}) can easily be classified into a specific aluminium alloy by considering the clearly recognizable characteristic peaks. The short-term measurement of 1s in (\subref{fig:plot_b}) is very noisy and statistical methods cannot recognize the characteristic peaks. Classifying alloys by such spectra is the goal of this paper.}} 
    \label{plot_Al_1}
\end{figure}

Prompt Gamma Neutron Activation Analysis (PGNAA) examines the integral metal composition without destruction.
It excites atoms by a neutron field, such that the de-excitation leads to measurable gamma radiation, which is characteristic for the specific element~\cite{Paul.2000}.
As you can see in Fig.~\ref{plot_Al_1}, we use this spectral data for each alloy.
For each measured energy level, we can see the measured amount of gamma radiation as "Counts".
The different elements of which the alloys consist cause peaks in the PGNAA spectra at different energy levels, by which it is possible to classify the different alloys using a long-term measurement.
However, short-term measurements result in very noisy spectra without clearly detectable peaks, as shown in Fig.~\ref{fig:plot_b}.

The paper \cite{Shayan.2023} shows that a classification of scrap metal with PGNAA is possible in short measurement times, based on the assumption of 50,000 counts per second for sampling.
There, the spectrum is viewed as a categorical probability distribution of the energy levels and this probability distribution can be estimated reliably by a long-term measurement.
Using this distribution, one can sample more spectral data to train machine learning algorithms and use the maximum likelihood method for real-time scrap metal classifications.

It remains open which gamma detectors are best suited for PGNAA-based scrap metal detection in real time and how machine learning methods can be improved for even quicker classification results.
Hence, this paper focuses on the optimization of metal recycling using new sensor solutions in combination with machine learning algorithms, with a specific focus on the differentiation of copper and aluminum alloys in real time.

We compare two different types of gamma detectors, one is a semiconductor detector equipped with a germanium crystal, called high purity germanium (HPGe) detector, the other a scintillator detector based on cerium bromide (CeBr$_{3}$).
Selecting the appropriate detector requires careful consideration of cost, measurement time, and accuracy.
While the HPGe detector is very expensive, the CeBr$_{3}$ detector is more affordable, but might have a worse performance.
The detectors measure a different amount of counts per second and have a different energy resolution.
In addition to the sampling methods in~\cite{Shayan.2023}, we test further methods for data generation (Section~\ref{data_generation_methods}) such as Conditional Variational Autoencoder (CVAE).
Furthermore, we evaluate various preprocessing methods like reducing and aggregating energy levels or weighting of different energy levels and classification methods like a Maximum Likelihood Classifier (MLC), XGBoost (XGB), Logistic Regression (LR) and a Linear Support Vector Machine (Linear SVM).

We achieved the best classification results with the MLC together with CVAE as data generation. 
When working with short measurement times, the CeBr$_{3}$ detector obtained better results than the HPGe detector. 
At high measurement times, this phenomenon occurred in reverse.

\section{Data and Detectors}
We analyse different metal alloys, which differ mainly in their content of aluminium or copper.
Two different detectors were used to identify the materials. On the one hand the high purity germanium detector (HPGe), where a very high energy resolution can be achieved when detecting gamma rays~\cite{HPGe} and on the other hand the cerium bromide detector (CeBr$_{3}$), which has a greater sensitivity and can be used with room-temperature~\cite{CeBr}. The HPGe detector has a resolution of 16,384 energy levels, while the CeBr$_{3}$ detector has a resolution of 2,048.
The different data available to us can be seen in Table~\ref{tab:data_overview}. 
For every material, we have spectra for five different alloys.

\begin{table}[h]
\centering
\caption{\emph{Overview of the recorded data for copper and aluminum in different states and with different types of detectors.}}
\label{tab:data_overview}
\begin{tabular}{cllcc}
\hline
\textbf{material} & \textbf{state} & \textbf{detector} & \textbf{counts per second} \\ \hline
copper            & block            & HPGe              & 30,000                      \\
aluminium         & block            & HPGe              & 19,000                      \\
aluminium         & chips            & HPGe              & 7,000                       \\
aluminium         & chips            & CeBr$_{3}$        & 11,000                      \\ \hline
\end{tabular}
\end{table}

Measurements were sometimes taken using metal in block form and other times using metal chips.
Metal blocks are solid pieces of metal formed into specific shapes or sizes, often used as raw materials or components in manufacturing processes. Metal chips refer to small pieces or shavings of metal generated during machining or cutting operations, typically considered waste material.
The chips have a lower density as they were measured in a cup with air between the different pieces. As the block only consist of the metal, the density and the weight of the metal are higher. 
This phenomenon can be seen in Table~\ref{tab:data_overview}, as the counts per second for block are significantly higher than for the chips.

\section{Methods}

In the following, we describe our approach to determine the best machine learning method for classifying the composition of metal scrap using different PGNAA methodologies. 
For most of the classification methods, we use the scikit-learn library~\cite{scikit-learn}, except for the neural network, where we use PyTorch~\cite{pytorch}. For data generation with a Conditional Variational Autoencoder (CVAE) and data preprocessing with a Denoising Autoencoder (DAE), we use TensorFlow~\cite{tensorflow}.

\subsection{Data Generation} \label{data_generation_methods}
We were able to achieve our results by using only one given long-term measured spectrum $S$ ($\ge 1h$) per alloy. We used these to generate short-term spectra $s$ ($< 10s$) for the training of our machine learning methods. This is an advantage, since real measurement data is expensive to produce and therefore only a limited number of measurements are available.
For this, we have two different approaches which are explained below.

\subsubsection{Categorical Sampling}
For this data generation method, we create a categorical probability distribution from the long-term measurement $S$. To obtain this distribution, we divide the counts $S=(c_1,\ldots,c_n)$ per energy level $(1, \ldots ,n)$ by the sum of all counts $(u = c_1 + \ldots + c_n)$. As a result of this, we get the relative frequencies $(\frac{c_1}{u}, ... , \frac{c_n}{u})$, which serve as an estimator for the categorical probability distribution $p(x|S)$ of a single gamma quantum of energy $x$. 
We use direct independent sampling from this distribution to generate test data, which closely replicates real-world conditions. To ensure training and test data diversity, the long-term measurements are divided into six shorter measurements by using dependent sampling. With the probability distributions generated as described above from these six measurements the short-term training spectra are generated. This enhances model robustness and mitigates overfitting risks.
A longer simulated measurement time leads to more frequent sampling from the long-term spectrum, which results in a higher number of simulated counts.

\subsubsection{Conditional Variational Autoencoder (CVAE)}
As an alternative for data generation, we use a CVAE, which is a modification of the Variational Autoencoder (VAE)~\cite{Kingma.20.12.2013}.
A VAE consists of a probabilistic encoder $q_\phi(z|s)$ and a probabilistic decoder $p_\theta(s|z)$, where $s$ is a (short-term) spectrum and $z$ is the latent representation. 
The probability distributions of the encoder and decoder are dependent on the parameters $\phi$ and $\theta$, which are learned during training.
The decoder can also be referred to as a generative model.
To generate artificial data, we sample latent variables from the prior $p(z)$ and feed them into the trained decoder $p_\theta(s|z)$, where we sample again.
With a CVAE~\cite{NIPS2015_8d55a249}, we additionally feed a label for the long-term spectrum $S$ to the decoder $p_\theta(s|z,S)$ in order to generate samples for a certain class.
For the encoder and decoder, we chose fully-connected neural networks with a single hidden layer, as in~\cite{Kingma.20.12.2013}.
For the training of the CVAE we use 10,000 sampled spectra.
When calculating the loss, we use a $\beta$-parameter~\cite{I.Higgins.2016} and set $\beta=\frac{N}{M}$ with $N$ being the size of the input $s$ and $M$ being the size of the latent space $z$.

\subsection{Preprocessing}

\subsubsection{Scaling}
Initial tests indicated that using the standard scaler is beneficial for the models XGBoost and Neural Network. For the other models, we use unscaled data. 

\subsubsection{Subsetting}\label{subsetting}
Subsetting involves the reduction and aggregation of energy levels by using just a subset of the spectra. 
An application could be the deletion of higher energy levels from the spectra.
This technique was applied on spectra obtained from the HPGe detector.

\subsubsection{Denoising Autoencoder (DAE)}
A DAE~\cite{Vincent.2008} is a modification of the traditional Autoencoder where the input is corrupted in order to achieve robustness to small changes in input.
We use a DAE to produce long-term measurements out of short-term measurements.
A classifier, which was trained using long-term measurements, should now be able to classify the output of the DAE.
In~\cite{Kim.2021}, a similar approach is used for angle-resolved photoemission spectroscopy data.
We use the fully-connected neural network architecture from~\cite{Dobilas.04.04.2022} and test different numbers of neurons in each layer.

\subsubsection{Escape and Double Escape Peaks (EPs and DEPs)}
EPs are characteristic features in gamma spectra that occur when measuring gamma rays. They arise due to interactions between the gamma rays and the detector material~\cite{EscapePeaks}. These are detection events where not all of the photon's energy is converted in the detector, but a certain, consistently sized portion of the energy is detected. If the peak at energy $E$ is above 1022 keV, an Escape Peak at $E - 511$ keV can occur in the spectrum~\cite{Metwally2005}. This results in an additional peak in the spectrum at a correspondingly lower energy. Similarly, a DEP can arise from an EP at $E - 1022$ keV.
As the location of the EPs and DEPs depend on the material, a different weighting of them can help classifying the different materials.
We only use this method with Maximum Likelihood Classifier (MLC), because here the reference spectrum can be weighted dependent on the EPs and DEPs of the alloy.

\subsubsection{Unique Peaks}
A possibility to improve the classification is the weighting of peaks, which only appear at one alloy. Therefore, we detected those peaks at the long-term measurements and accordingly weighted the different training and test spectra.

\subsection{Classification} \label{classification_methods}
We did each test five times and then reported the average to counteract randomness.
Unless otherwise stated, we used 10,000 spectra for training.
In order to achieve sufficient representativeness, we worked with 5,000 test spectra each.

\subsubsection{Maximum Likelihood Classifier (MLC)}
With MLC~\cite{millar2011maximum} we compare the categorical probability distribution from the short-term measurement to the different long-term measurements to find the best fitting distribution. 
To classify the short-term measurement $s$, take the number of short-term photons measured $s=(c_1',\ldots,c_n')$ per energy level and multiply it by the appropriate log relative probability of the long-term measurement $S$ to obtain the log-likelihood: 

\begin{equation*}
\log(p(s|S)) = \sum_{i=1}^{n}{c_i' \cdot \log\left(\frac{c_i + 1}{\sum_{i=1}^{n}{(c_i+1)}}\right).}
\end{equation*}

Since there were difficulties in the case when $c_i = 0$, we add the value 1 to each number of photons of the long-term measurements. 

Instead of just using the distribution of the long-term measurement, we use measurements generated by the Categorical Sampling and vary time and amount of reference distributions. In case of multiple distributions per alloy, the mean of the log-likelihoods is taken and compared to classify the short-term measurement $s$. The best results were achieved using 500 reference spectra per alloy, each corresponding to a measurement time of 1,800 seconds. 

\subsubsection{XGBoost (XGB)}
For its capacity to handle large datasets and model complex patterns, we selected XGBoost, a robust ensemble classifier using
gradient boosting~\cite{chen2016xgboost}. Initial tests favor its linear-based~\cite{xgb-docs} variant over tree-based methods in our context, hence adopting it exclusively for our purpose.

\subsubsection {Logistic Regression (LR)}
It is a statistical approach for modeling binary outcomes based on predictor variables. It extends linear regression to the logistic function, making it suitable for categorical predictions. 
The training algorithm uses the one-vs-rest (OvR) scheme for a multiclass case.

\subsubsection{Linear Support Vector Machine (Linear SVM)}
The SVM obtained the best results with a linear kernel.
Therefore, we used the linear SVM to improve the runtime~\cite{Suthaharan2016}.

\subsubsection{K-nearest Neighbors (KNN)}
The KNN algorithm is based on the nearest neighbor decision rule proposed by Cover and Hart~\cite{cover1967nearest}. It operates on the principle of identifying the nearest neighbor(s) of an example and assigning the example to the same class as its nearest neighbor(s).

\subsubsection{Radius Neighbors Classifier (RNC)}
The RNC, an extension of the nearest neighbor decision rule~\cite{cover1967nearest}, classifies examples by considering all data points within a specified radius. By assessing the proximity of points within the defined radius, the algorithm assigns a class label based on local patterns.

\subsubsection{Neural Network (NN)}
We used the PyTorch library~\cite{pytorch} for the implementation of a fully connected, 4-layer network. The exact structure is described in more detail in the appendix~\ref{Hyperparameters}. 

\subsubsection{Random Forest Classifier (RFC)}
The RFC~\cite{breiman2001random} is an ensemble learning method that combines multiple decision trees to make predictions.

\subsubsection{Extra Trees Classifier (ETC)}
The ETC~\cite{geurts2006extremely} builds an ensemble of decision trees with an extra level of randomness in comparison to the RFC. This randomness is used to reduce over-fitting and speeding up training~\cite{geurts2006extremely}.

\subsubsection{Kuiper test (Kui)}
It is a statistical method used to classify and compare probability distributions.
Kui takes into account the largest positive and negative differences between the cumulative distribution functions of two samples.
This was implemented using basic Pandas operations. One short-term measurement is compared with all long-term measurements. We determine the smallest absolute maximum difference between the distributions of long-term and short-term measurements respective energy levels in order to assign the metal alloy. The smallest difference is identified as the most probable assignment for the short-term measurement.

\section{Results}
\subsection{Classification}

We tested the classifiers from Section~\ref{classification_methods} with various parameters using Categorical Sampling as data generation and the HPGe detector with block data.
As you can see in Table~\ref{tab:all_methods}, the best results could be achieved with MLC, XGB, LR and Linear SVM.

\begin{table}[H]
\begin{center} 
\caption{\label{tab:all_methods} \emph{Accuracy of all examined models on aluminum and copper block measurements with Categorical Sampling as data generation, HPGe detector and MLC achieving the best results}. }
\begin{tabular}{lrrrrrr}
\toprule
classifier & \multicolumn{3}{c}{aluminium} & \multicolumn{3}{c}{copper} \\
 & \multicolumn{3}{c}{measurement time} & \multicolumn{3}{c}{measurement time} \\
 & 0.5s & 1.0s & 2.0s & 0.5s & 1.0s & 2.0s\\
\midrule
MLC & \textbf{96.32} & \textbf{99.03} & \textbf{99.87} & \textbf{91.87} & \textbf{95.18} & \textbf{97.96}\\
XGB & 93.93 & 98.19 & 99.65 & 89.42 & 93.48 & 97.09\\
LR & 93.28 & 97.69 & 99.50 & 88.82 & 93.00 & 96.51\\
Linear SVM & 91.94 & 97.28 & 99.55 & 88.21 & 92.43 & 96.24\\
NN & 90.80 & 95.62 & 98.45 & 87.51 & 91.85 & 95.22\\
KNN & 89.18 & 95.41 & 98.63 & 86.57 & 91.44 & 95.06\\
RFC & 88.58 & 95.34 & 98.66 & 85.21 & 90.01 & 94.20\\
ETC & 89.94 & 95.34 & 98.63 & 85.90 & 90.48 & 94.66\\
RNC & 89.03 & 95.26 & 98.66 & 86.46 & 91.51 & 95.06\\
Kui & 77.07 & 84.32 & 91.53 & 83.68 & 89.56 & 94.29\\
\bottomrule
\end{tabular}

\vspace{.2cm}
\end{center}
\end{table}

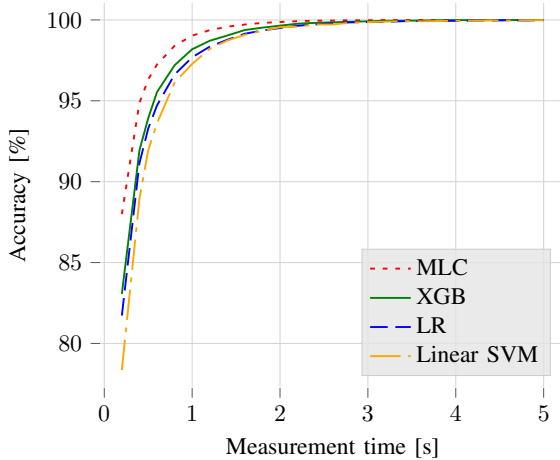
\begin{figure}[H]
\begin{center}
\begin{tikzpicture}[scale=0.9]

\definecolor{dimgray85}{RGB}{85,85,85}
\definecolor{gainsboro229}{RGB}{229,229,229}
\definecolor{green}{RGB}{0,128,0}
\definecolor{lightgray}{RGB}{211,211,211}
\definecolor{lightgray204}{RGB}{204,204,204}
\definecolor{orange}{RGB}{255,165,0}

\begin{axis}[
axis line style={white},
legend cell align={left},
legend style={
  fill opacity=0.8,
  draw opacity=1,
  text opacity=1,
  at={(0.97,0.03)},
  anchor=south east,
  draw=lightgray204,
  fill=gainsboro229
},
tick align=outside,
tick pos=left,
x grid style={lightgray},
xlabel=\textcolor{black}{Measurement time [s]},
xmajorgrids,
xmin=-0.0399999968707562, xmax=5.23999999985099,
xtick style={color=dimgray85},
y grid style={lightgray},
ylabel=\textcolor{black}{Accuracy [\%]},
ymajorgrids,
ymin=77.27128, ymax=101.08232,
ytick style={color=dimgray85}
]
\addplot [thick, red, dash pattern=on 2pt off 3.3pt]
table {%
0.200000002980232 87.9936
0.400000005960464 94.8704
0.5 96.3208
0.600000023841858 97.248
0.800000011920929 98.4
1 99.0296
1.20000004768372 99.3656
1.39999997615814 99.5808
1.60000002384186 99.7168
1.79999995231628 99.8168
2 99.8712
2.20000004768372 99.924
2.40000009536743 99.9536
2.59999990463257 99.9608
2.79999995231628 99.976
3 99.9816
3.20000004768372 99.9928
3.40000009536743 99.9864
3.59999990463257 99.996
3.79999995231628 99.996
4 99.9976
4.19999980926514 99.9976
4.40000009536743 99.9992
4.59999990463257 99.9992
4.80000019073486 100
5 100
};
\addlegendentry{MLC}
\addplot [thick, green]
table {%
0.200000002980232 83.0592
0.400000005960464 91.9304
0.5 93.9264
0.600000023841858 95.5376
0.800000011920929 97.1976
1 98.1864
1.20000004768372 98.7112
1.39999997615814 99.0376
1.60000002384186 99.3768
1.79999995231628 99.5256
2 99.6472
2.20000004768372 99.768
2.40000009536743 99.8192
2.59999990463257 99.8512
2.79999995231628 99.8856
3 99.916
3.20000004768372 99.948
3.40000009536743 99.9544
3.59999990463257 99.9616
3.79999995231628 99.9824
4 99.9888
4.19999980926514 99.9848
4.40000009536743 99.9896
4.59999990463257 99.992
4.80000019073486 99.9936
5 99.9896
};
\addlegendentry{XGB}
\addplot [thick, blue, dash pattern=on 7.4pt off 3.2pt]
table {%
0.200000002980232 81.7248
0.400000005960464 91.1376
0.5 93.2848
0.600000023841858 94.7136
0.800000011920929 96.6232
1 97.688
1.20000004768372 98.3488
1.39999997615814 98.772
1.60000002384186 99.1472
1.79999995231628 99.3488
2 99.4984
2.20000004768372 99.6392
2.40000009536743 99.7208
2.59999990463257 99.7992
2.79999995231628 99.8488
3 99.8704
3.20000004768372 99.88
3.40000009536743 99.9208
3.59999990463257 99.9312
3.79999995231628 99.9408
4 99.944
4.19999980926514 99.9648
4.40000009536743 99.9656
4.59999990463257 99.9784
4.80000019073486 99.9648
5 99.9848
};
\addlegendentry{LR}
\addplot [thick, orange, dash pattern=on 12.8pt off 3.2pt on 2pt off 3.2pt]
table {%
0.200000002980232 78.3536
0.400000005960464 88.9384
0.5 91.9352
0.600000023841858 93.6176
0.800000011920929 96.1264
1 97.2784
1.20000004768372 98.212
1.39999997615814 98.728
1.60000002384186 99.0832
1.79999995231628 99.324
2 99.5472
2.20000004768372 99.5784
2.40000009536743 99.7192
2.59999990463257 99.7184
2.79999995231628 99.8304
3 99.8688
3.20000004768372 99.8944
3.40000009536743 99.9304
3.59999990463257 99.9128
3.79999995231628 99.9344
4 99.9528
4.19999980926514 99.9568
4.40000009536743 99.9728
4.59999990463257 99.9696
4.80000019073486 99.9848
5 99.9848
};
\addlegendentry{Linear SVM}
\end{axis}

\end{tikzpicture}

\end{center}
\captionsetup{type=figure}
\captionof{figure}{\emph{Accuracy of the best models on aluminium measurements (Table~\ref{tab:all_methods})}.}
\label{fig:Best_classifiers_alu}
\end{figure}

\begin{figure}[H]
\begin{center}
\begin{tikzpicture}[scale=0.9]

\definecolor{dimgray85}{RGB}{85,85,85}
\definecolor{gainsboro229}{RGB}{229,229,229}
\definecolor{green}{RGB}{0,128,0}
\definecolor{lightgray}{RGB}{211,211,211}
\definecolor{lightgray204}{RGB}{204,204,204}
\definecolor{orange}{RGB}{255,165,0}

\begin{axis}[
axis line style={white},
legend cell align={left},
legend style={
  fill opacity=0.8,
  draw opacity=1,
  text opacity=1,
  at={(0.97,0.03)},
  anchor=south east,
  draw=lightgray204,
  fill=gainsboro229
},
tick align=outside,
tick pos=left,
x grid style={lightgray},
xlabel=\textcolor{black}{Measurement time [s]},
xmajorgrids,
xmin=-0.0399999968707562, xmax=5.23999999985099,
xtick style={color=dimgray85},
y grid style={lightgray},
ylabel=\textcolor{black}{Accuracy [\%]},
ymajorgrids,
ymin=79.89736, ymax=100.72344,
ytick style={color=dimgray85}
]
\addplot [thick, red, dash pattern=on 2pt off 3.3pt]
table {%
0.200000002980232 87.0896
0.400000005960464 90.7512
0.5 91.8712
0.600000023841858 92.708
0.800000011920929 94.1512
1 95.18
1.20000004768372 96.008
1.39999997615814 96.6608
1.60000002384186 97.2368
1.79999995231628 97.6312
2 97.9584
2.20000004768372 98.3184
2.40000009536743 98.5632
2.59999990463257 98.7424
2.79999995231628 98.9024
3 99.1168
3.20000004768372 99.2112
3.40000009536743 99.2816
3.59999990463257 99.4224
3.79999995231628 99.4512
4 99.5648
4.19999980926514 99.6328
4.40000009536743 99.664
4.59999990463257 99.704
4.80000019073486 99.7256
5 99.7768
};
\addlegendentry{MLC}
\addplot [thick, green]
table {%
0.200000002980232 83.2376
0.400000005960464 87.8464
0.5 89.4232
0.600000023841858 90.3568
0.800000011920929 91.8696
1 93.4832
1.20000004768372 94.4576
1.39999997615814 95.1792
1.60000002384186 96.0104
1.79999995231628 96.5664
2 97.0864
2.20000004768372 97.4904
2.40000009536743 97.8392
2.59999990463257 98.148
2.79999995231628 98.4016
3 98.5968
3.20000004768372 98.7464
3.40000009536743 98.888
3.59999990463257 99.0544
3.79999995231628 99.1976
4 99.3144
4.19999980926514 99.3656
4.40000009536743 99.4
4.59999990463257 99.5208
4.80000019073486 99.5344
5 99.6104
};
\addlegendentry{XGB}
\addplot [thick, blue, dash pattern=on 7.4pt off 3.2pt]
table {%
0.200000002980232 82.8624
0.400000005960464 87.5528
0.5 88.8192
0.600000023841858 89.8752
0.800000011920929 91.6568
1 93.0016
1.20000004768372 93.8512
1.39999997615814 94.7496
1.60000002384186 95.4688
1.79999995231628 96.0416
2 96.5056
2.20000004768372 96.936
2.40000009536743 97.3552
2.59999990463257 97.5768
2.79999995231628 97.8992
3 98.1376
3.20000004768372 98.2936
3.40000009536743 98.4064
3.59999990463257 98.64
3.79999995231628 98.6968
4 98.8672
4.19999980926514 99.0496
4.40000009536743 99.0408
4.59999990463257 99.168
4.80000019073486 99.2272
5 99.3048
};
\addlegendentry{LR}
\addplot [thick, orange, dash pattern=on 12.8pt off 3.2pt on 2pt off 3.2pt]
table {%
0.200000002980232 80.844
0.400000005960464 86.8832
0.5 88.2072
0.600000023841858 89.332
0.800000011920929 91.2896
1 92.4264
1.20000004768372 94.0784
1.39999997615814 94.7712
1.60000002384186 95.664
1.79999995231628 95.524
2 96.2448
2.20000004768372 97.1256
2.40000009536743 97.4088
2.59999990463257 97.492
2.79999995231628 98.0208
3 98.2056
3.20000004768372 98.3432
3.40000009536743 98.5072
3.59999990463257 97.9808
3.79999995231628 98.7704
4 98.9696
4.19999980926514 98.884
4.40000009536743 99.1456
4.59999990463257 99.1648
4.80000019073486 99.3208
5 99.1264
};
\addlegendentry{Linear SVM}
\end{axis}

\end{tikzpicture}
\end{center}
\captionsetup{type=figure}
\captionof{figure}{\emph{Accuracy of the best models on copper measurements (Table~\ref{tab:all_methods})}.}
\label{fig:Best_classifiers_cu}
\end{figure}
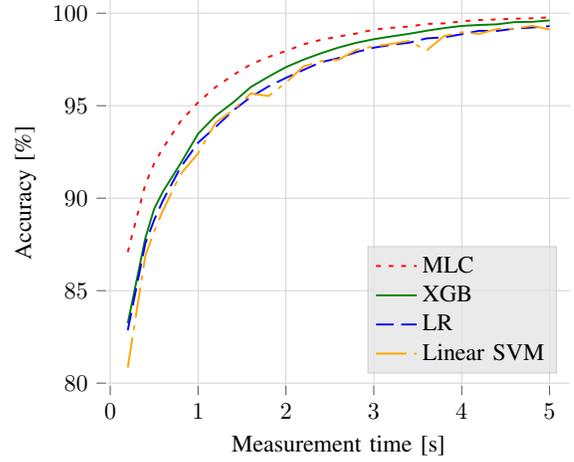
To see the differences of the four best classifiers, you can see more detailed plots in Fig.~\ref{fig:Best_classifiers_alu} and Fig.~\ref{fig:Best_classifiers_cu} for aluminium and copper. In the following experiments, we will further examine the results of these classifiers using different data generation and preprocessing methods, as well as different types of detectors.

\subsection{Data Generation}
Table~\ref{tab:cvae_alu} shows the results of the four best classifiers with CVAE data generation as well as the increase or decrease in accuracy compared to Categorical Sampling data generation. CVAE led to a slight increase in the accuracy of MLC while the accuracy of the other classifiers decreased. 
Thus, in the following experiments, we use MLC in combination with CVAE and the other classifiers in combination with Categorical Sampling.

\begin{table}[H]
\begin{center} 
\caption{\label{tab:cvae_alu} \emph{Accuracy of best models on aluminum and copper block measurements with CVAE data generation and HPGe detector. Only MLC improves using the CVAE data generation}.}
\begin{tabular}{lrrr}
\toprule
classifier & \multicolumn{3}{c}{measurement time} \\
aluminium  & 0.5s & 1.0s & 2.0s \\
\midrule
MLC & \textbf{96.52} (+0.20) & \textbf{99.12} (+0.09) & \textbf{99.91} (+0.04) \\
LR & 86.37 (-6.91) & 94.09 (-3.60) & 98.32 (-1.18) \\
Linear SVM & 84.38 (-7.56) & 93.32 (-3.96) & 98.10 (-1.45) \\
XGB & 51.61 (-42.32) & 62.67 (-35.52) & 76.14 (-23.51) \\
\midrule
copper  & 0.5s & 1.0s & 2.0s \\
\midrule
MLC & \textbf{92.09} (+0.22) & \textbf{95.27} (+0.09) & \textbf{98.07} (+0.11) \\
LR & 81.59 (-7.23) & 89.19 (-3.81) & 94.30 (-2.21) \\
Linear SVM & 77.04 (-11.17) & 87.12 (-5.31) & 93.62 (-2.62) \\
XGB & 51.21 (-38.21) & 60.92 (-32.56) & 73.78 (-23.31) \\
\bottomrule
\end{tabular}

\vspace{.2cm}
\end{center}
\end{table}

\subsection{Preprocessing}
In Table~\ref{tab:sub}, a subset of the entire spectra was selected as described in Section~\ref{subsetting}. 
For each, training and testing, a dataset of 1,000 spectra per material with a simulated measurement time of one second were created via Categorical Sampling and CVAE. The number of energy levels describe the upper boundary from which on features were deleted. A reduction in the number of features diminishes the overall performance of the respective model. However, this reduction does not exhibit a linear relationship with the number of features. This observation suggests that there is not a substantial information gain in the last energy levels of the spectra. It is the case because the first energy levels have a considerably higher count-value. In the higher energy levels, oftentimes no count is measured, resulting in no information gain. The initial energy levels are considerably more crucial, proving sufficient for achieving a relatively high level of accuracy. However, because using all the features leads to the overall best results, and the prediction runtime is very quick in any case, we decided to keep the original energy level count for our further experiments.

\begin{table}[t]
\begin{center} 
\caption{\label{tab:sub} \emph{Accuracy decrease of the best models with decreased amount of energy levels for aluminium and copper blocks with HPGe detector and a simulated measurement time of one second}.}
\begin{tabular}{lrrrrrr}
\toprule
classifier & \multicolumn{3}{c}{aluminium} & \multicolumn{3}{c}{copper}\\
& \multicolumn{3}{c}{number of energy levels} & \multicolumn{3}{c}{number of energy levels}\\
  & 4000 & 8000 & 16384 & 4000 & 8000 & 16384\\
\midrule
MLC & \textbf{98.29} & \textbf{98.69} & \textbf{99.04} & \textbf{93.83} & \textbf{94.23} & \textbf{95.32}\\
LR & 96.65 & 97.66 & 97.93 & 91.40 & 91.74 & 92.69\\
Linear SVM & 96.04 & 96.98 & 97.38 & 91.24 & 92.37 & 92.87\\
XGB & 97.03 & 97.47 & 98.08 & 92.73 & 92.52 & 93.35\\
\bottomrule
\end{tabular}

\vspace{.2cm}
\end{center}
\end{table}

Using a DAE to convert the short term measurements to 60 seconds measurements and classifying those led to a decrease in accuracy. For aluminium, the accuracy decreased by approx. 6\% and for copper by approx. 7\% on average for the best four classifiers respectively.

The preprocessing with EPs and DEPs was only tested with MLC and mostly brings small improvements for the accuracy. The results can be seen in Table~\ref{tab:escape_peaks}. 
For these experiments, we used 1,000 spectra for training and five long-term measurements as reference spectra.

The preprocessing with Unique Peaks brings slight positive and negative variations in the accuracy.
In total, it is difficult to say whether the variations in the accuracy are due to the randomness of the data or due to the preprocessing with Unique and Escape Peaks. 

\begin{table}[H]
\begin{center} 
\caption{\label{tab:escape_peaks} \emph{Accuracy of MLC with CVAE as data generation, HPGe detector, block data and EPs and DEPs as preprocessing}.}
\begin{tabular}{lrrr}
\toprule
        material & \multicolumn{3}{c}{measurement time} \\
                   &           0.5s &  1.0s &  2.0s \\
\midrule
aluminium &          96.67 (+0.15) & 99.15 (+0.03) & 99.92 (+0.01)\\
copper &          91.78 (-0.31)& 95.39 (+0.12)& 98.28 (+0.21)\\
\bottomrule
\end{tabular}

\vspace{.2cm}
\end{center}
\end{table}

\subsection{Detectors}
In our analysis so far, we have evaluated the accuracy of four top classifiers - MLC, XGB, Linear SVM and LR - using aluminium and copper data in block form, with the MLC achieving the best results.
In this section of the investigation, we turn to aluminium chips data, as only data in this form is available for the CeBr$_{3}$ detector.
In order to examine the results, we will restrict ourselves to the MLC in the following, as it shows better results compared to other methods. 
With a measurement time of 0.5 seconds, the CeBr$_{3}$ detector has a maximum accuracy of 62.58\%, which is significantly higher than HPGe detector with 50.40\% using Categorical Sampling to generate the data. When using CVAE, the accuracy barely increases, 62.88\% for CeBr$_{3}$ detector and 50.69\% for HPGe detector.

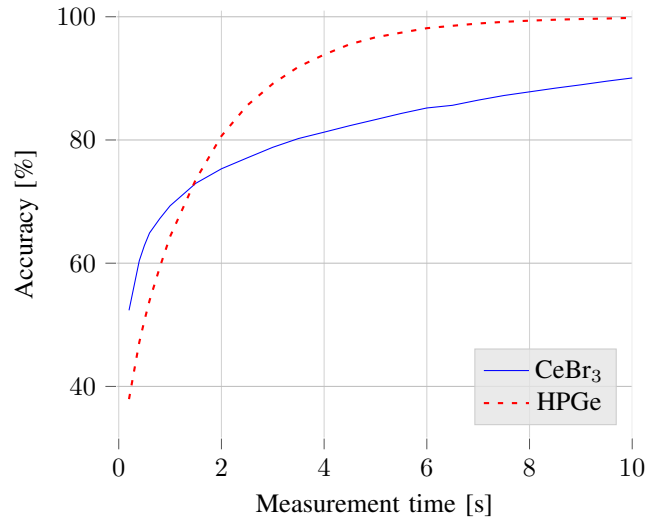
\begin{figure}[h]
\begin{center}
\begin{tikzpicture}

\definecolor{dimgray85}{RGB}{85,85,85}
\definecolor{gainsboro229}{RGB}{229,229,229}
\definecolor{green}{RGB}{0,128,0}
\definecolor{lightgray}{RGB}{211,211,211}
\definecolor{lightgray204}{RGB}{204,204,204}
\definecolor{orange}{RGB}{255,165,0}

\begin{axis}[
    axis line style={white},
    legend style={
        fill opacity=0.8,
        draw opacity=1,
        text opacity=1,
        at={(0.97,0.03)},
        anchor=south east,
        draw=lightgray204,
        fill=gainsboro229
    },
    tick align=outside,
    tick pos=left,
    x grid style={lightgray},
    xlabel=\textcolor{black}{Measurement time [s]},
    xmajorgrids,
    xmin=-0.0399999968707562, xmax=10,
    xtick style={color=dimgray85},
    ylabel=\textcolor{black}{Accuracy [\%]},
    ymajorgrids,
    ymax=101.08232,
    ytick style={color=dimgray85}
]

\addplot[color=blue, mark=none]
    coordinates {
        (0.2, 52.352)
        (0.4, 60.3984)
        (0.5, 62.8808)
        (0.6, 64.9104)
        (0.8, 67.244)
        (1.0, 69.316)
        (1.5, 72.9632)
        (2.0, 75.3224)
        (2.5, 77.0848)
        (3.0, 78.81199999999999)
        (3.5, 80.23279999999999)
        (4.0, 81.268)
        (4.5, 82.3256)
        (5.0, 83.30719999999999)
        (5.5, 84.3024)
        (6.0, 85.19439999999999)
        (6.5, 85.6264)
        (7.0, 86.4664)
        (7.5, 87.21439999999999)
        (8.0, 87.8128)
        (8.5, 88.4088)
        (9.0, 88.95200000000001)
        (9.5, 89.5376)
        (10.0, 90.0648)
    };
    \addlegendentry{CeBr$_{3}$}
\addplot [thick, red, dash pattern=on 2pt off 3.3pt]
    coordinates {
        (0.2, 37.938400000000005)
        (0.4, 47.160799999999997)
        (0.5, 50.6912)
        (0.6, 53.9368)
        (0.8, 59.35119999999999)
        (1.0, 64.32)
        (1.5, 73.5608)
        (2.0, 80.64)
        (2.5, 85.5984)
        (3.0, 89.1288)
        (3.5, 91.8768)
        (4.0, 93.8536)
        (4.5, 95.53839999999999)
        (5.0, 96.6624)
        (5.5, 97.416)
        (6.0, 98.1424)
        (6.5, 98.5328)
        (7.0, 98.90320000000001)
        (7.5, 99.1744)
        (8.0, 99.3648)
        (8.5, 99.52)
        (9.0, 99.62799999999999)
        (9.5, 99.712)
        (10.0, 99.8232)
    };
    \addlegendentry{HPGe}

\end{axis}

\end{tikzpicture}

\end{center}
\captionsetup{type=figure}
\captionof{figure}{\emph{Comparison of the accuracy over time of the two detectors with aluminium chips data using CVAE and MLC.}}
\label{fig:Best_classifiers_different_detectors}
\end{figure}

In Fig.~\ref{fig:Best_classifiers_different_detectors} we have determined the accuracy as a function of time from 0.2 to 10 seconds measurement time in steps of 0.2 seconds.
While the CeBr$_{3}$ delivers the best results for measurement times up to 1.3 seconds, the HPGe achieves better accuracy for longer measurement times.
Both detectors converge to 100\% over time, with HPGe converging much faster.
The analysis of the results shows that the HPGe detector only needs 10 seconds to reliably distinguish all aluminium alloys, while the CeBr$_{3}$ detector takes significantly longer for the same task. 
As shown in the corresponding figure, the curve for the HPGe detector ends at 10 seconds, at which point a classification accuracy of 100\% is achieved. 
In contrast, the CeBr$_{3}$ detector achieves full classification accuracy of 100\% only after about 80 seconds.


\section{Discussion}
As the Linear SVM and Logistic Regression achieve good results on our data, it seems that the data is linear seperable in a high dimension.
In addition to the linear separability of the spectra, the regularization techniques provided by XGBoost in the form of L1 and L2 are the reason why the classifier generalizes well and thus achieves very good accuracy.

The good results of MLC could be attributed to its ability to model the data distribution more accurately. 
In scenarios where the underlying data distribution is complex and doesn't conform strictly to linear relationships, like our spectral data, MLC can capture the nuances better than the other classifiers. 

\begin{figure}[t]
\begin{center}
\input{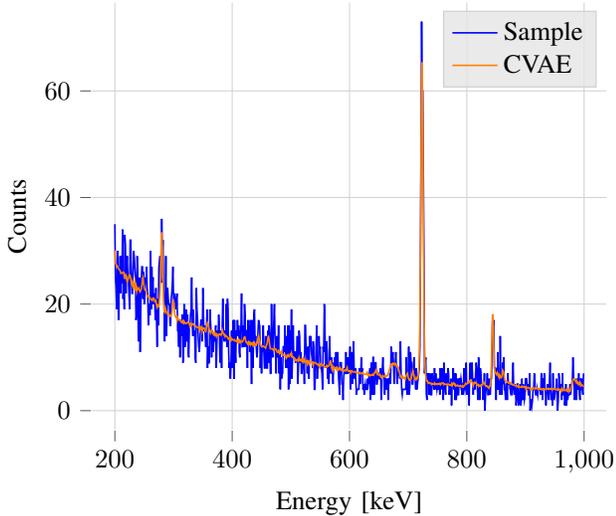}
\end{center}
\captionsetup{type=figure}
\captionof{figure}{\emph{Section of an example spectrum of an aluminium alloy generated with different methods}.}
\label{fig:data_gen_spectrum}
\end{figure}

Using CVAE as data generation, we see that the results of MLC can be improved while for the other classifiers, the accuracy decreases. 
A reason could be that the CVAE generated spectra contain less noise and thus are more suitable as a probability distribution for MLC.
A comparison of the spectra generated by the different methods is shown in Fig.~\ref{fig:data_gen_spectrum}.
Thanks to this characteristic, MLC requires less CVAE spectra than spectra generated by Categorical Sampling to achieve good results, because for the latter it needs more data to build a mean representing the data adequately.
This reduces the prediction runtime and thus allows a faster classification.
For the other classifiers, the accuracy decreases using CVAE as data generation.
This could be due to the fact that the CVAE data is not as similar to the test data in its variance as the Categorical Sampling data. Thus, the Categorical Sampling data is more suitable for training.

The CeBr$_{3}$ data results are better than the HPGe data results, because the higher number of counts at short measurement times has a greater influence on the determination due to a higher probability of detector response. 
This is important when measuring very low levels of gamma radiation.

The high resolution of the HPGe detector leads to fewer counts per channel when the same number of counts is distributed across more channels,
increasing relative uncertainty of each count rate. 
This makes it more challenging to identify significant patterns between the aluminium alloys.

The results from the chip data are worse than the block data because the mass of the sample is significantly reduced and therefore fewer nuclei are irradiated and fewer gamma rays are counted.

The HPGe detector shows its strengths at longer measurement times, which could be due to its very high resolution to precisely detect even fine energy differences.
Peaks carrying relevant information regarding classification are thus emphasised more clearly.
Otherwise, the number of counts in the CeBr$_{3}$ is not sufficient at longer measurement times for the compensation that would be necessary to distinguish the aluminium alloys.

\section{Conclusion}

In conclusion, the research demonstrates the potential of applying advanced classification methods and data generation techniques to improve metal alloy differentiation in recycling processes based on PGNAA measurements.
The MLC was found to be the most effective classification method combined with CVAE data generation technique.

The study shows that the CeBr$_{3}$ detector excels in scenarios requiring very short measurement times, but overall the
HPGe detector generally remains the preferable choice  to deliver accurate results.
The choice of detector depends on the specific needs of the application.

Because real short-term measurement data is lacking, it is necessary to perform and study short-term measurements in detail in the future to verify the applicability of our findings in the real world.

\appendix
\section*{Hyperparameters} \label{Hyperparameters}

\subsection{Data Generation}

\begin{table}[H]
\centering
\begin{tabular}{p{4.5cm}|p{2.5cm}}
\hline
\textbf{Conditional Variational Autoencoder} & \textbf{Value} \\
\hline
library & TensorFlow \\
scaling & MinMaxScaler\footnotemark[1] \\
hidden\_units & 100 \\
latent\_variables & 10 \\
loss & $-\text{ELBO}$\footnotemark[2] \\ 
optimizer & Adam\footnotemark[3] \\
learning\_rate & 0.001 \\
batch\_size & 32 \\
epochs & 100 \\
\hline
\end{tabular}
\end{table}

\footnotetext[1]{implementation of the scikit-learn~\cite{scikit-learn} library}
\footnotetext[2]{Negative ELBO as in~\cite{Kingma.20.12.2013} using MSE, a $\beta$-parameter and KL divergence }
\footnotetext[3]{Adam optimization algorithm~\cite{Kingma.22.12.2014}}

\subsection{Preprocessing}

\begin{table}[H]
\centering
\begin{tabular}{p{4.5cm}|p{2.5cm}}
\hline
\textbf{Denoising Autoencoder} & \textbf{Value} \\
\hline
library & TensorFlow \\
scaling & MinMaxScaler\footnotemark[1] \\
hidden\_units & 10 \\
latent\_variables & 100 \\
loss & MeanSquaredError \\
optimizer & Adam\footnotemark[3] \\
learning\_rate & 0.001 \\
batch\_size & 32 \\
epochs & 100 \\
\hline
\end{tabular}
\end{table}

\newpage
\subsection{Classification}

%

\begin{table}[H]
\centering
\begin{tabular}{p{5cm}|p{2cm}}
\hline
\textbf{XGBoost} & \textbf{Value} \\
\hline
booster                         &       gblinear \\
device                          &       cpu \\
verbosity                       &       1 \\
lambda                          &       0.25 \\
alpha                           &       0 \\
updater                         &       shotgun \\
feature\_selector                &       cyclic \\
learning\_rate                   &       0.45 \\
\hline
\end{tabular}
\end{table}
%

\begin{table}[H]
\centering
\begin{tabular}{p{5cm}|p{2cm}}
\hline
\textbf{Logistic Regression} & \textbf{Value} \\
\hline
penalty                     &       L2 \\
dual                        &       False \\
tol                         &       0.0001 \\
C                           &       1.0 \\
fit\_intercept              &       True \\
intercept\_scaling          &       1 \\
class\_weight               &       None \\
solver                      &       lbfgs \\
max\_iter                   &       150 \\
multi\_class                 &       auto \\
verbose                     &       0 \\
\hline
\end{tabular}
\end{table}
%
\begin{table}[H]
\centering
\begin{tabular}{p{5cm}|p{2cm}}
\hline
\textbf{Linear Support Vector Machine} & \textbf{Value} \\
\hline
penalty                     &       L2 \\
loss                        &       squared\_hinge \\
dual                        &       True \\
tol                         &       1 \\
C                           &       3 \\
multi\_class                &       ovr \\
fit\_intercept              &       True \\
intercept\_scaling          &       1 \\
class\_weight               &       None \\
verbose                     &       0 \\
random\_state               &       None \\
max\_iter                   &       100 \\
\hline
\end{tabular}
\end{table}
%
\begin{table}[H]
\centering
\begin{tabular}{p{5cm}|p{2cm}}
\hline
\textbf{K-nearest Neighbors} & \textbf{Value} \\
\hline
n\_neighbors                 &       8000 \\
weights                     &       distance \\
metric                      &       euclidean \\
algorithm                   &       brute \\
leaf\_size                   &       30 \\
p                           &       2 \\
\hline
\end{tabular}
\end{table}
%
\begin{table}[H]
\centering
\begin{tabular}{p{5cm}|p{2cm}}
\hline
\textbf{Radius Neighbors Classifier} & \textbf{Value} \\
\hline
radius                       &       500 \\
weights                      &       distance \\
metric                       &       euclidean \\
algorithm                    &       brute \\
leaf\_size                   &       30 \\
p                           &       2 \\
outlier\_label               &       most\_frequent \\

\hline
\end{tabular}
\end{table}
\begin{table}[H]
\centering
\begin{tabular}{p{2.7cm}|p{2cm}|p{2cm}}
\hline
\textbf{Neural Network} & \textbf{Value Blocks} & \textbf{Value Chips} \\
\hline
library                        &       \multicolumn{2}{c}{PyTorch}                 \\
n\_layers                        &       \multicolumn{2}{c}{4}                 \\
layer                           &       \multicolumn{2}{c}{fully connected} \\
activation\_function             &       \multicolumn{2}{c}{relu}                \\
criterion                       &       \multicolumn{2}{c}{CrossEntropyLoss} \\
optimizer                       &       \multicolumn{2}{c}{Adam\footnotemark[3]}                \\
num\_epochs                      &       2                   &   5 \\
n\_batch\_size                    &       500                 &   1000\\
learning\_rate                   &       \multicolumn{2}{c}{0.0001}   \\
\hline
\end{tabular}
\end{table}

\footnotetext[3]{Adam optimization algorithm~\cite{Kingma.22.12.2014}}
%
\begin{table}[H]
\centering
\begin{tabular}{p{5cm}|p{2cm}}
\hline
\textbf{Random Forest Classifier} & \textbf{Value} \\
\hline
n\_estimators                 &       4000 \\
min\_samples\_split            &       50 \\
criterion                    &       gini \\
max\_depth                    &       None \\
min\_samples\_leaf                &    1 \\
min\_weight\_fraction\_leaf        &   0.0 \\
max\_features                    &   sqrt \\
max\_leaf\_nodes                  &   None \\
min\_impurity\_decrease           &   0.0 \\
bootstrap                       &   True \\
oob\_score                       &   False \\
max\_samples                     &   None \\
\hline
\end{tabular}
\end{table}
\begin{table}[H]
\centering
\begin{tabular}{p{5cm}|p{2cm}}
\hline
\textbf{Extra Trees Classifier} & \textbf{Value} \\
\hline
n\_estimators                  &       4000 \\
min\_samples\_split             &       50 \\
criterion                     &       entropy \\
max\_depth & None \\
min\_samples\_leaf & 1 \\
max\_features & sqrt \\
min\_weight\_fraction\_leaf       & 0.0 \\
max\_leaf\_nodes                & None \\
min\_impurity\_decrease         & 0.0 \\
bootstrap                       & False \\
verbose                     &       0 \\
max\_samples                     &   None \\
\hline
\end{tabular}
\end{table}

\section*{Acknowledgment}

Research was conducted as part of the MetalClass project, funded by the German Federal Ministry of Education and Research with the funding reference
number 01IS20082A/B.
The authors express their sincere gratitude to our project partner AiNT GmbH for their support and enriching discussions throughout the project's duration.
Some of the research was conducted while Kai Krycki was with AiNT GmbH.
Special acknowledgment is also extended to Wieland-Werke AG for graciously providing the copper alloys essential to our investigation.

\bibliographystyle{IEEEtran}
\bibliography{7.literature}


\end{document}